\documentclass[conference]{IEEEtran}
\usepackage{cvpr}
\usepackage{times}
\usepackage{epsfig}
\usepackage{graphicx}
\usepackage{amsmath}
\usepackage{amssymb}
\usepackage{authblk}
\usepackage{caption}
\usepackage{subcaption}

\usepackage{graphicx}
\usepackage{caption}
\usepackage{subcaption}
\usepackage{amsmath}
\usepackage{setspace}
\usepackage{algorithm}
\usepackage[english]{babel}
\usepackage[utf8]{inputenc}
\usepackage{algorithm}
\usepackage[noend]{algpseudocode}

\usepackage{epstopdf}
\usepackage{amssymb}
\usepackage{authblk}
\usepackage{diagbox}
\usepackage{threeparttable}
\usepackage{graphicx}
\usepackage{booktabs}
\usepackage{chngcntr}
\usepackage{multirow}
\usepackage{cite}
\usepackage{float}
\usepackage{threeparttable}
\usepackage{amssymb}
\usepackage{caption}

\usepackage[breaklinks=true,bookmarks=false]{hyperref}

\cvprfinalcopy 

\def\cvprPaperID{****} 

\begin{document}

\title{Long Term Prediction Of Neural Activity Using LSTM With Multiple Temporal Resolution }

\author[1]{Yilin Song}
\author[1]{Ran Wang}
\author[1]{Yao Wang }
\author[2]{Jonathan Viventi }

\affil[1]{Department of Electrical and Computer Engineering,  New York University, NY, USA}
\affil[2]{Department of Biomedical Engineering, Duke University, Durham, NC, USA}


\maketitle

\begin{abstract}


Epileptic seizures are caused by abnormal, overly synchronized, electrical activity in the brain. The abnormal electrical activity manifests as waves, propagating across the brain. Accurate prediction of the propagation velocity and direction of these waves could enable real-time responsive brain stimulation to suppress or prevent the seizures entirely. However, this problem is very challenging because the algorithm must be able to predict the neural signals in a sufficiently long time horizon to allow enough time for medical intervention. We consider how to accomplish long term prediction using a LSTM network. To alleviate the vanishing gradient problem, we propose two encoder-decoder-predictor structures, both using multi-resolution representation. The novel LSTM structure with multi-resolution layers could significantly outperform the single-resolution benchmark with similar number of parameters. To overcome the blurring effect associated with video prediction in the pixel domain using standard mean square error (MSE) loss, we use energy-based adversarial training to improve the long-term prediction. We demonstrate and analyze how a discriminative model with an encoder-decoder structure using 3D CNN model improves long term prediction. 

\end{abstract}

\section{Introduction}
Studies have focused on seizure prediction for decades, but reliable prediction of seizure activity many minutes before a seizure has been elusive. Constructed features like wavelet, energy of spike and spectral power \cite{bandarabadi2015epileptic,li2013seizure,eftekhar2014ngram,gadhoumi2012discriminating, netoff2009seizure, chua2009automatic,sorensen2010automatic,temko2011eeg,acharya2011automatic} are applied on electroencephalogram (EEG) or electrocorticographic (ECoG) data with coarse resolution for most of current neurological analysis works. Prior work has focused on predicting seizures minutes or hours in advance of the seizure, using supervised datasets with labeled examples of seizures. However, with the rich spatial and temporal patterns unveiled by high resolution micro-electrocorticographic ($\mu$ECoG) \cite{viventi2011flexible} which is very similar to a high-frame rate video signal, accurate prediction of neural activities at the sub-second level could become a very interesting and tractable problem. Neural signal prediction on this time frame would allow responsive stimulation to suppress seizures. This kind of neural signal prediction could also find a compact representation for neural activity which could lead to understanding non-pathologic neural activity. To capture the highly non-linear dynamics in neural activities, deep learning neural networks appear to be a promising solution. But, learning a compact representation for neural video prediction gets more challenging when trying to predict in a longer future and the deep neural network model develops a more severe vanishing gradient problem.

To model long term dependencies, Long Short Term Memory (LSTM) units \cite{hochreiter1997long} were proposed as an improvement over vanilla RNN to solve vanishing gradients problem by introducing gate functions. Gated Recurrent Unit (GRU) \cite{cho2014learning} as simplified version of LSTM units has achieved better performance in a number of applications \cite{kaiser2015neural, trischler2016natural,bahdanau2014neural}. Even though LSTM and GRU tries to solve vanishing gradient problem by preserve long term dependency in their cells, modeling long term dependencies is still difficult. For neural language translation, instead of decoding a sequence from a compact feature learnt through an encoding network, \cite{bahdanau2014neural,rocktaschel2015reasoning} use word-by-word attention mechanism which allow direct connection between premise and hypothesis sentences. By using such direct connections, it alleviates the vanishing gradient problem for long sentence translation. Another different approach is to use memory augmented neural networks as Neural Turing Machines \cite{graves2014neural}. By using external memory to store information, the explicit storage of hidden states creates a shortcut through time. \cite{santoro2016one, gulcehre2017memory} both achieve good performance by using external memory network. 

\begin{figure*}
        \centering
       	\begin{subfigure}[b]{0.98\textwidth}
	
		\includegraphics[width=\textwidth]{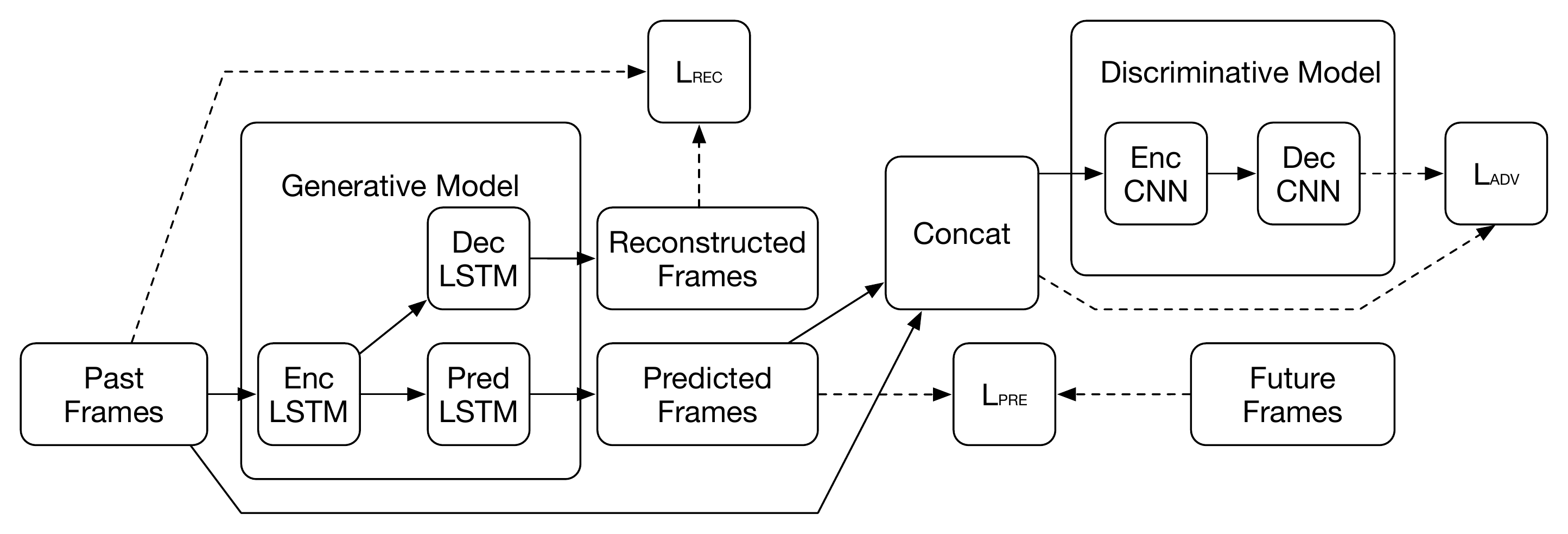}
        \end{subfigure}
         \caption{Video prediction framework. The generative model is built using convolutional LSTM \cite{xingjian2015convolutional}. The network flow is represented with solid arrow, whereas the losses for the generative model are represented with dashed arrows.  }
\label{framework_fig}
\end{figure*} 

Because convolution account for short-range dependencies, to capture long term correlation, CNN based models have to either increase the depth of the model or use a larger receptive field and larger stride. Either way is generally considered not optimal in time series prediction. Work in \cite{srivastava2015highway, he2016deep} creates efficient information flow from lower layer to high layer by using residual modules and skip connections. \cite{oord2016pixel} used residual module between depth layer in both CNN and RNN based image pixel prediction. Because time series such as audio and video have high temporal correlation, to increase the receptive field with same number of parameters \cite{van2016wavenet} used diluted convolution kernel and achieve good performance.
Inspired by diluted convolution in \cite{van2016wavenet}, we propose a LSTM network that uses multi-resolution layers. The higher layer skips each temporal connections to 
create a shortcut, while lower layer is temporally connected and preserves the fine grained information. We also experiment with an explicitly multi-resolution LSTM structure that resembles a temporal pyramid. We demonstrate both multi-resolution representations improve long term prediction compare to a benchmark LSTM.

Learning long term dependencies not only needs an appropriate network structure but also needs a suitable loss function. For video prediction, to overcome blurry predictions caused by using pixel-wise mean square error (MSE), \cite{mathieu2015deep} added total variation loss. Video prediction could be consider as a special case of domain transfer, where the past observed frames lies on one data manifold and future frame lies on another one. Adversarial training finds the relationship between these two manifolds. \cite{lotter2015unsupervised, mathieu2015deep} add adversarial loss on top of MSE. But how video prediction benefits from adversarial training are not fully understood. To further understand how adversarial training benefits video prediction, we use a encoder-decoder 3D CNN structure for the discriminative model. The discriminative model uses reconstruction error as its loss rather than KL-divergence measure. This resembles energy-based GAN in \cite{zhao2016energy} versus GAN \cite{goodfellow2014generative}.

%
%
%

\section{Framework}

In this section, we describe the general structure of our neural video prediction model. The structure consists of two different models, a generative model and a discriminative model. The generative model first takes past observations of video sequences as input and learns a compact feature representation, from which the generative model then reconstructs the past frames and predicts future frames. We explore different model structures during the experiments, and use convolutional LSTM \cite{xingjian2015convolutional} as basic building block for generative model. The discriminative model structure is nearly the same in all experiments. Its main goal is to determine whether the future frames are generated conditioned on the true past frames. Together with the generative model, these two models are considered as adversarial training \cite{goodfellow2014generative}. The general structure is shown in Fig.~\ref{framework_fig}.

\subsection{Generative Model}  
Let $X = \{x_1 \cdots x_{t+n} \}$ denotes a video sequence, where $x_t$ denotes current observation, $x_{t+n}$ denotes the $n$th frame in the future to predict. The generative network takes $\{x_1 \cdots x_{t}\}$ as input and outputs a sequence $Y =\{y_1 \cdots y_{t+n} \}$. 

In our approach the generative model has an encoder network, a decoder network and a predictor network similar as \cite{srivastava2015unsupervised}. These networks all use convolutional LSTM \cite{xingjian2015convolutional} as basic computation module. The encoder network takes $\{x_1 \cdots x_t\}$ as input and generates a representation $l$. The decoder network reconstructs $\{y_1 \cdots y_t\}$ from $R_{x_1,\cdots,x_t}$. The decoder LSTM is set to be a conditional model namely the decoder reconstructs $y_{t-m}$ from $y_{t-m+1}$. The predictor network generates $\{y_{t+1} \cdots y_{t+n}\}$ from $R_{x_1,\cdots,x_t}$. The predictor model is also a conditional model and it conditions on $y_{t+m}$ to predict $y_{t+m+1}$.  The loss for the generative model consists four parts:
\begin{equation}
\begin{aligned}
\mathcal{L}_G &= \lambda_{rec} \mathcal{L}_{rec} + \lambda_{pred}\mathcal{L}_{pred} +\lambda_{adv} \mathcal{L}_{adv}   \\
\mathcal{L}_{rec} &= \sum_{i=1}^t || x_i - y_i||^2_2  \\
\mathcal{L}_{pred} &= \sum_{i={t+1}}^{t+n}|| x_i - y_i||^2_2  \\
\mathcal{L}_{adv} &= ||Dec(Enc(Z)) - Z||^2_2 \\
\end{aligned}
\label{total_loss}
\end{equation}

Z is a four dimensional tensor of size $c \times (t+n) \times h \times w$, with $c$, $h$ and $w$ represents channel, height and width of the frame respectively. Z is constructed by stacking $\{x_1 \cdots x_{t}\}$ and $\{y_{t+1} \cdots y_{t+n}\}$ in time order. $\mathcal{L}_{rec}$, $\mathcal{L}_{pred}$ are the pixel domain loss for the reconstructed frames and predicted frames respectively. Whereas $\mathcal{L}_{adv}$ is the adversarial loss from the discriminative model. 

\begin{figure*}
        \centering
	
		\includegraphics[width=0.95\textwidth]{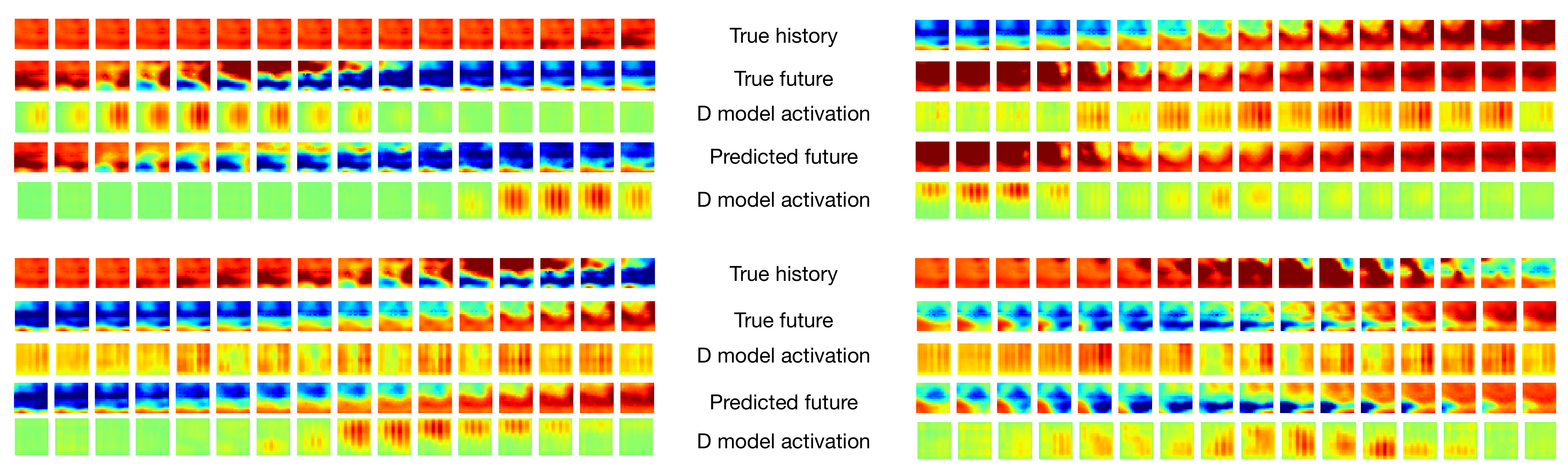}
		\caption{Understanding the benefit from adversarial training: The input to the discriminative model is either true history with true future or true history with predicted future. Third and fifth row of each example shows the activation of second to last layer output across all channel. The activation by true data is distributed almost evenly in both space and time domain to reconstruct the entire sequence. The activation by the sequence with predicted future however concentrates on spatial and temporal inconsistencies. For example, in the first sequence, the discriminative model finds the inconsistency in the last few frames.}
     
\label{model_understanding}
\end{figure*} 
\subsection{Discriminative Model}
\begin{figure}
        \centering
        	\begin{subfigure}[b]{0.45\textwidth}
		\includegraphics[width=\textwidth]{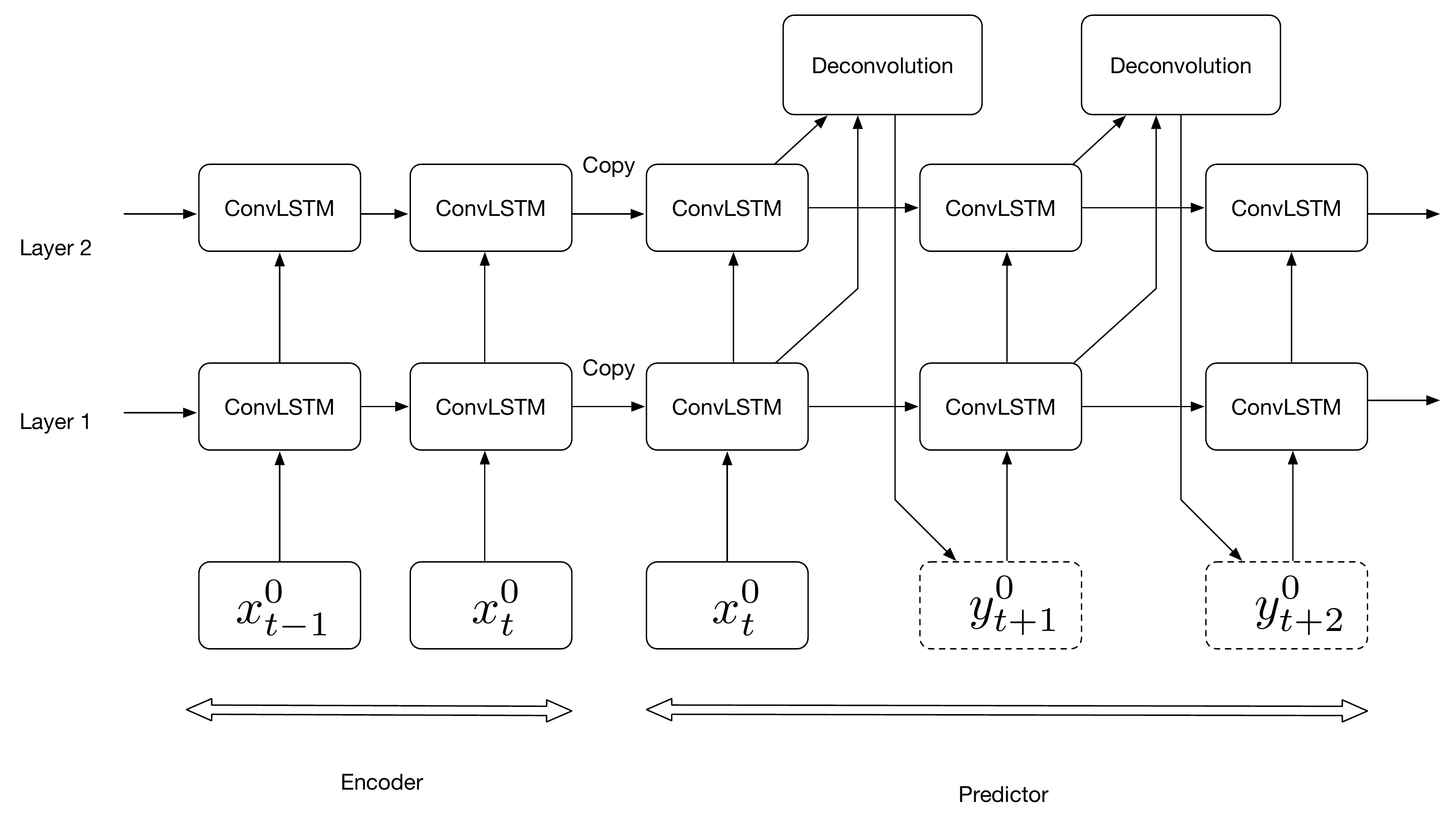}
		\caption{benchmark network}
        \end{subfigure}      
        
       	\begin{subfigure}[b]{0.45\textwidth}
	
		\includegraphics[width=\textwidth]{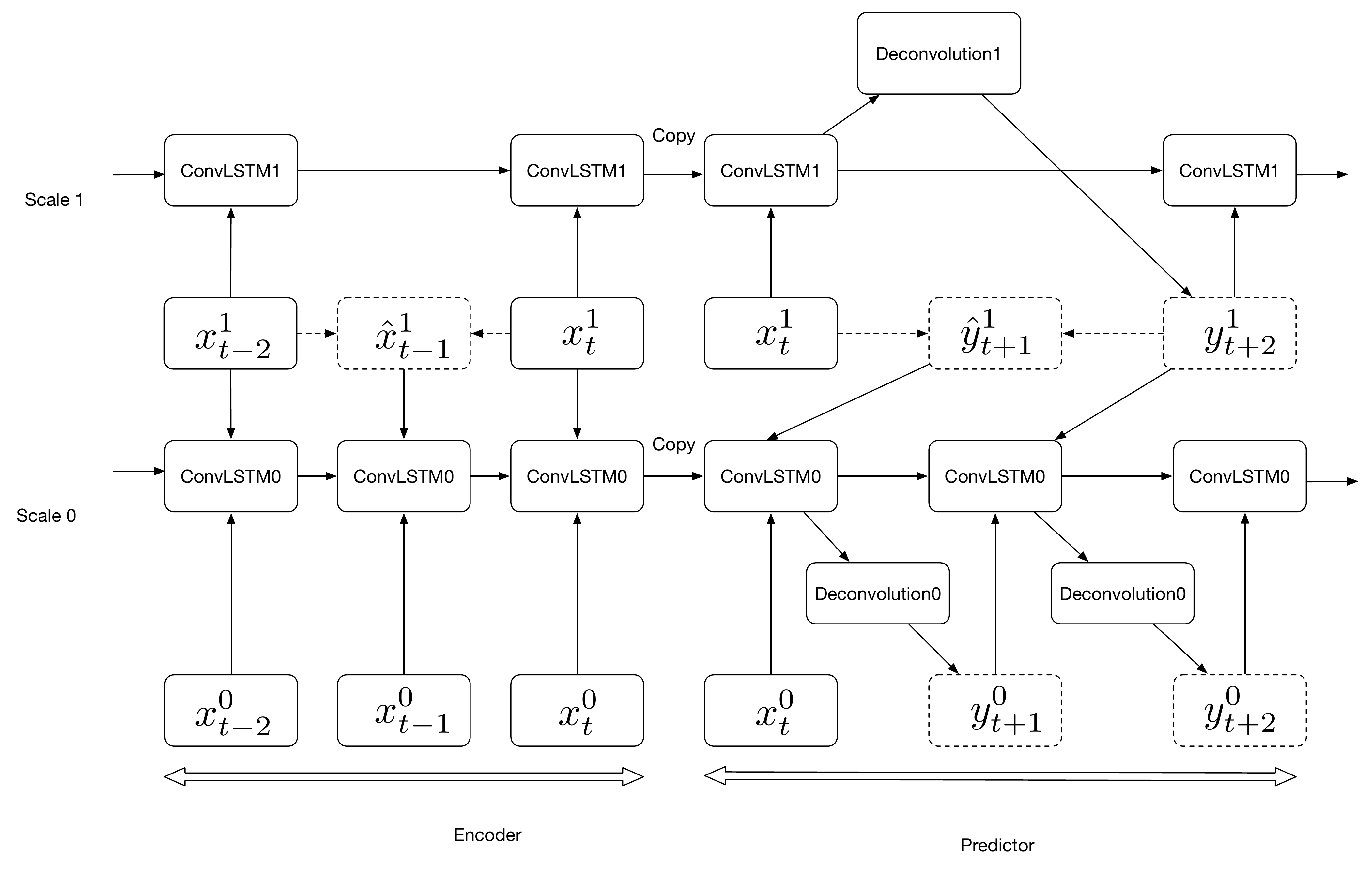}
		\caption{multi-resolution LSTM}
        \end{subfigure}
        
       	\begin{subfigure}[b]{0.45\textwidth}
	
		\includegraphics[width=\textwidth]{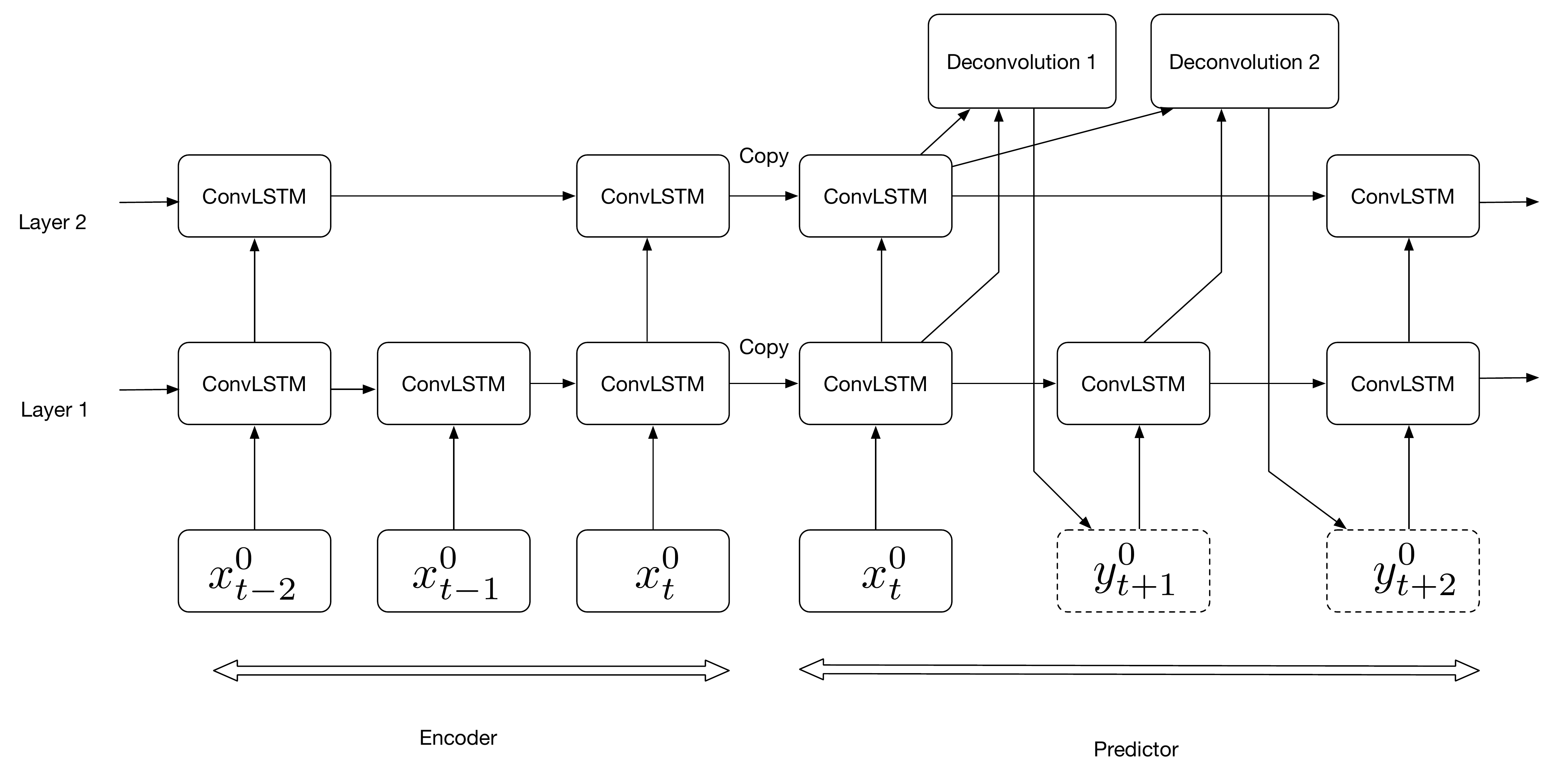}
		\caption{LSTM with multi-resolution layers}
        \end{subfigure}        
         	  
          \caption{Benchmark network, multi-resolution LSTM and LSTM with multi-resolution layers. The multi-resolution LSTM has two scales, and in each scale it has two layer structure. Only one layer is drawn per scale for simplicity. The dotted box represents the predicted frames and $\hat{y}$ represents linear interpolated frames.}


\label{framework}
\end{figure} 

Generative adversarial network(GAN) were introduced by \cite{goodfellow2014generative}, where image is generated from random noise by using two networks trained in a competing manner. The discriminative model in \cite{goodfellow2014generative} minimize the KL-divergence between true image distribution and generated image distribution. The original GAN structure suffers from convergence problem and collapsing mode \cite{salimans2016improved,arjovsky2017wasserstein}. To solve those problems, \cite{salimans2016improved} introduce several techniques including feature matching, minibatch discrimination and historical averaging. \cite{zhao2016energy} use image reconstruction loss instead of KL-divergence loss. \cite{radford2015unsupervised} derive a stable deep convolutional GAN structure by modifying modules in both generator and discriminator. \cite{arjovsky2017wasserstein} show the true data distribution and generative data distribution manifolds in high dimensional space hardly have any overlap, and Wasserstein distance is better compare to other distance measures for non-overlapping distribution. \cite{arjovsky2017wasserstein} achieved the state of art performance for image generation.   

For adversarial training for domain transfer problem, where one generates a sample in target domain condition on the data in source domain. In domain transfer unlike GAN, whose generative model could easily suffer from mode collapsing \cite{salimans2016improved, radford2015unsupervised}, overlaps between source domain and target domain manifold is easier to find. \cite{ledig2016photo, taigman2016unsupervised,lotter2015unsupervised,mathieu2015deep} all used modest model structure to perform domain transfer task and achieved good performance. Video prediction could be considered as a domain transfer problem, where the past frames embedding lies on one manifold and future embedding lies on another manifold. \cite{lotter2015unsupervised} concatenates the LSTM features of past frames and CNN feature of generated frame to train a separate multilayer perceptron. \cite{mathieu2015deep} uses a multi-scale 2d convolutional network, the discriminative model stacks all input frames in the channel dimension and output a single scalar indicating whether the video frames are generated or from ground truth future. But both networks fail to model the temporal correlation between frames explicitly.


More importantly, it is not fully understood how adversarial training benefits video prediction. To exploit the temporal dependancies, we use an auto encoder and decoder 3D CNN structure as our discriminative model. The discriminative model uses the energy as the loss function. Energy-based model finds compact representation for the sequence which lives on a low dimension manifold. \cite{zhao2016energy} demonstrate the energy-based GAN training has advantage over GAN for image generation.  Another benefit of using encoder-decoder structure is by mapping the activation into pixel space, it helps understanding how adversarial training benefits video prediction. Figure~\ref{model_understanding} shows the activation in the second to last layer of the discriminative model when provided different input. 
The loss for discriminative model is:
\begin{equation}
\begin{aligned}
\mathcal{L}_{D} =&||Dec(Enc(X)) - X||^2_2  - \\
	&||Dec(Enc(Z)) - Z||^2_2
\end{aligned}
\label{D_loss}
\end{equation}

The Dec and Enc in Eq.~\ref{D_loss} refers to the encoder and decoder in the discriminative model.



\section{Benchmark and Multi-resolution Network}
For neural video prediction, to capture the long term dependencies, we propose two different network structures: multi-resolution LSTM and LSTM with multi-resolution layers. For all generative models, the basic building block is ConvLSTM \cite{xingjian2015convolutional}. Each ConvLSTM layer at each time takes $X_t$ as input, and has memory cell state $C_t$, hidden state $H_t$ and gates $i_t, f_t, o_t$. The equation we use for ConvLSTM are shown in Eq.~\ref{ConvLSTM}, where $\ast$ denotes convolution operator and $\circ$ denotes Hadamard product. For all generative models, they all have an encoder, a decoder and a predictor. 
\begin{equation}
\begin{aligned}
i_t &= \sigma ( W_{xi} \ast {X}_t + W_{hi} \ast H_{t-1} + W_{ci} \circ C_{t-1} + b_i) \\
f_t &= \sigma ( W_{xf}  \ast {X}_t + W_{fi}  \ast H_{t-1} +W_{cf} \circ C_{t-1}+b_f) \\
C_t &= f_t\circ C_{t-1}+i_t \circ tanh(W_{xc}\ast X_t+ W_{hc}\ast H_{t-1}+b_c) \\
o_t &= \sigma(W_{xo}  \ast {X}_t + W_{ho}  \ast H_{t-1} +W_{co }\circ C_{t}+b_o) \\
H_t &= o_t \circ tanh(C_t) \\
\end{aligned}
\label{ConvLSTM}
\end{equation}

\subsection{Benchmark Network}
First we introduce the benchmark ConvLSTM model, which is a two layer ConvLSTM structure shown in Fig.~\ref{framework}(a). In the benchmark model, each convolution LSTM layer uses a convolution kernel of size $5 \times 5$. For the predictor and decoder convLSTM in the model, the output of both layers go through a deconvolution layer. The deconvolution layer uses kernel size of $1\times 1$ and outputs a frame, which is essentially a weighted average of all input feature maps followed by a $tanh$ function.

\subsection{Multi-resolution LSTM}
In this approach, in general, we generate $k$ temporal scales of the training sequences. The original sequence constitutes  scale 0,  and the upper scales are recursively down-sampled from the lower scale by a factor of 2. The top scale (coarsest resolution) works in the same way as the benchmark network over its samples only.  The lower scale considers both the samples in that scale as well as the interpolated samples from the upper scale.  We only present the 2-scale case here for simplicity.  In order to avoid delay,  we use the simple averaging of the current and the previous sample in the lower scale as the anti-aliasing filter for downsampling.  Specifically let $x_i^0$ represents the true video frame at time $i$.  Scale 1 signal at even time samples is produced by:
\begin{equation}
x^1_{i} = \frac{ x^0_{i-1} + x^0_{i}}{2},i=\text{even}
\end{equation}
To interpolate the odd samples at scale 1 from even samples, we use simple averaging interpoation filter. The interpolated signal from scale 1 is
\[
u(x^1)(i)= 
\begin{cases}
x^1_i \quad i= \text{even} \\
\frac{ x^1_{i-1} + x^1_{i+1}}{2}, i =\text{odd} \\
\end{cases} \\
\]

\begin{figure*}
        \centering
       	\begin{subfigure}[b]{0.9\textwidth}
	
		\includegraphics[width=\textwidth]{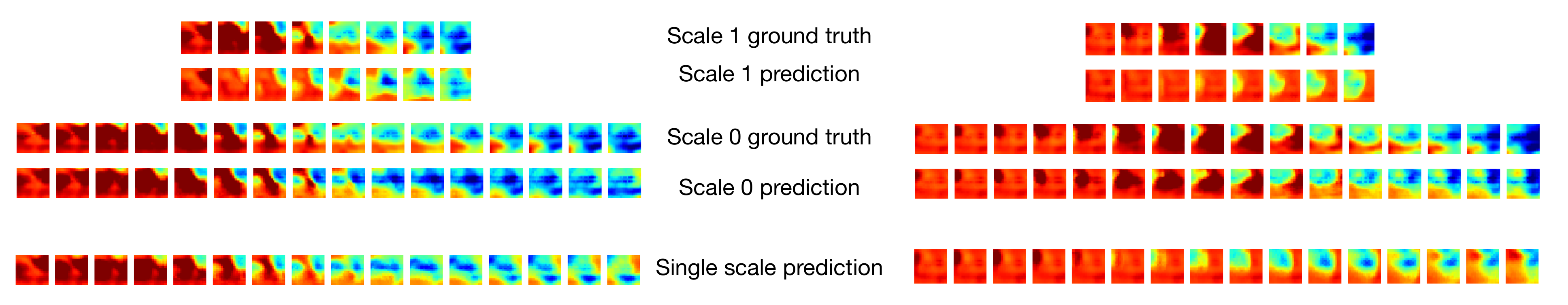}
        \end{subfigure}
        \caption{Comparison between 2-scale and single scale model for video prediction. For single scale video prediction, the encoder, decoder and predictor in the generative model each uses two layer convolutional LSTM. In 2-scale prediction, for each scale the model have the same network structure as the single scale benchmark. The single scale model and multi-resolution LSTM each corresponds to model 6 and 7 in Tab.~\ref{PSNR_compare}.}

\label{temporal_pyramid_seq}
\end{figure*}

As shown in Fig.~\ref{framework}(b), we first predict even samples at scale 1. We then interpolate the odd future samples from the predicted even samples, to generate all predicted samples, $u(y^1)(t+i)$, from scale 1.  We then predict samples at scale 0 using both past samples at scale 0 and current predicted sample at scale 1. Specifically, we predict  samples at $t+i$ using the features learned up to time $t+i-1$ (from both scales), the actual or predicted sample at $t+i-1$ at scale 0, as well as the predicted sample at $t+i$ at scale 1, i.e., $y^0_{i+i} = G(f_{t+i-1}, y^0_{t+i-1}, u(y^1)(t+i))$.  The two inputs $y^0_{t+i-1}$ and  $u(y^1)(t+i)$ to the ConvLSTM predictor are simply stacked as two channels at the same time. In each scale, the generative model is trained by minimizing the loss function at that scale $(k=0, 1)$:

\begin{equation}
\begin{aligned}
\mathcal{L}_G^k &= \lambda_{rec}^k \mathcal{L}_{rec}^k + \lambda_{pred}^k\mathcal{L}_{pred}^k +\lambda_{adv}^k \mathcal{L}_{adv}^k   \\
\mathcal{L}_{rec}^k &= \sum_{i \in \text{scale} \, k} ||x^k_i - y_i^k||_2^2 \\
\mathcal{L}_{pred}^k &= \sum_{i \in \text{scale} \, k} ||x^k_{t+i} - y_{t+i}^k||_2^2 \\
\mathcal{L}_{adv}^k &= ||Dec(Enc(Z^k)) - Z^k||^2_2 \\
\end{aligned}
\label{scale_loss}
\end{equation} 
where $Z^k$ is a four dimensional tensor by stacking the true past frames and predicted frames for scale $k$ in time order. The illustration of multi-scale structure is shown in Fig.~\ref{framework}(b). In each scale, the LSTM network has exactly the same two-layer ConvLSTM structure as the benchmark model, except the input frame of scale 0 have twice the number of channels compared to scale 1: half from the current scale and another half from the upper scale. The comparison of 2-scale multi-resolution LSTM and single scale prediction are shown in Fig.~\ref{temporal_pyramid_seq}.

%

\subsection{LSTM with Multi-resolution layer}

\begin{figure}
        \centering
	
		\includegraphics[width=0.45\textwidth]{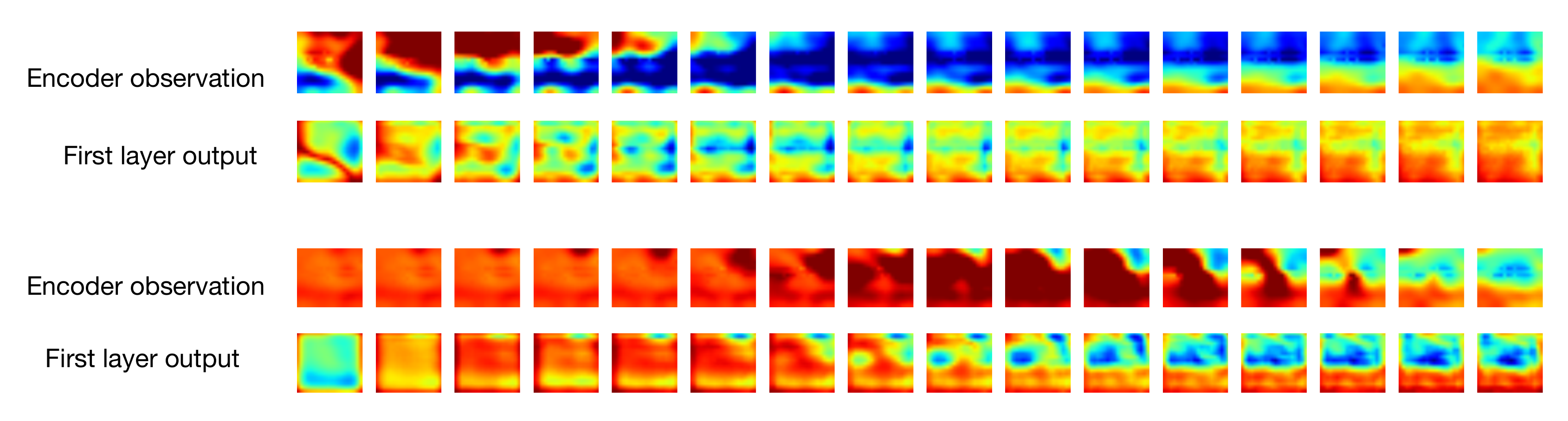}
		\caption{Demonstration of the high correlation of the first layer output of the two-layer LSTM model at the original resolution. Each layer used 2 layer convolutional LSTM each with 128 convolutional kernels with receptive field of $5 \times 5$. Row 1 and 3 show the encoder observation for two different sequences. Row 2 and 4 show the 
average of the 128 feature maps produced by layer 1. }
\label{temporal_highway_display}
\end{figure}

For time series prediction because the temporal correlation is high, in order to achieve a larger receptive field with the same amount of parameters, \cite{van2016wavenet} uses diluted convolution where the convolutional filter in higher layers of the CNN network are structured with zero coefficients every other connections. 

Inspired by \cite{van2016wavenet}, we propose a LSTM network that has multi-resolution layers. The network have a higher layer and a lower layer.  The fine-grained temporal resolution is preserved by the lower layer shown in in Fig.~\ref{temporal_highway_display}. The higher layer of the convolutional LSTM model use a skip temporal connection shown in Fig.~\ref{framework}(c). Compare to the lower layer, the higher layer creates a temporal highway, which alleviates the vanishing gradient problem. Different from the benchmark network shown in Fig.~\ref{framework}(a), the deconvolution layer in multi-resolution layer network (Fig.~\ref{framework}(c)) use different parameters to predict. In our implementation, the deconvolution layer is performing $1\times1$ spacial convolution on the feature map outputs, and the increase number of parameter compare to Fig.~\ref{framework}(a) is almost negelectable. The number of parameters used in different models are shown in Tab.~\ref{PSNR_compare}.



\section{Experiment}
\subsection{Dataset}
We analyzed $\mu$ECoG data from an acute in vivo feline model of seizures. The 18 by 20 array of high-density active electrodes has 500 $\mu$m spacing between nearby channels. The in vivo recording has a temporal sampling rate of 277.78 Hz and lasts 53 minutes. We obtained a total of 894K frames. In total, there are 788 K consecutive training frames and 106K consecutive testing frames. During training, we use 16 frames as observation to predict the next 16 frames.


\subsection{Results}

\begin{figure}
        \centering
	
		\includegraphics[width=0.5\textwidth]{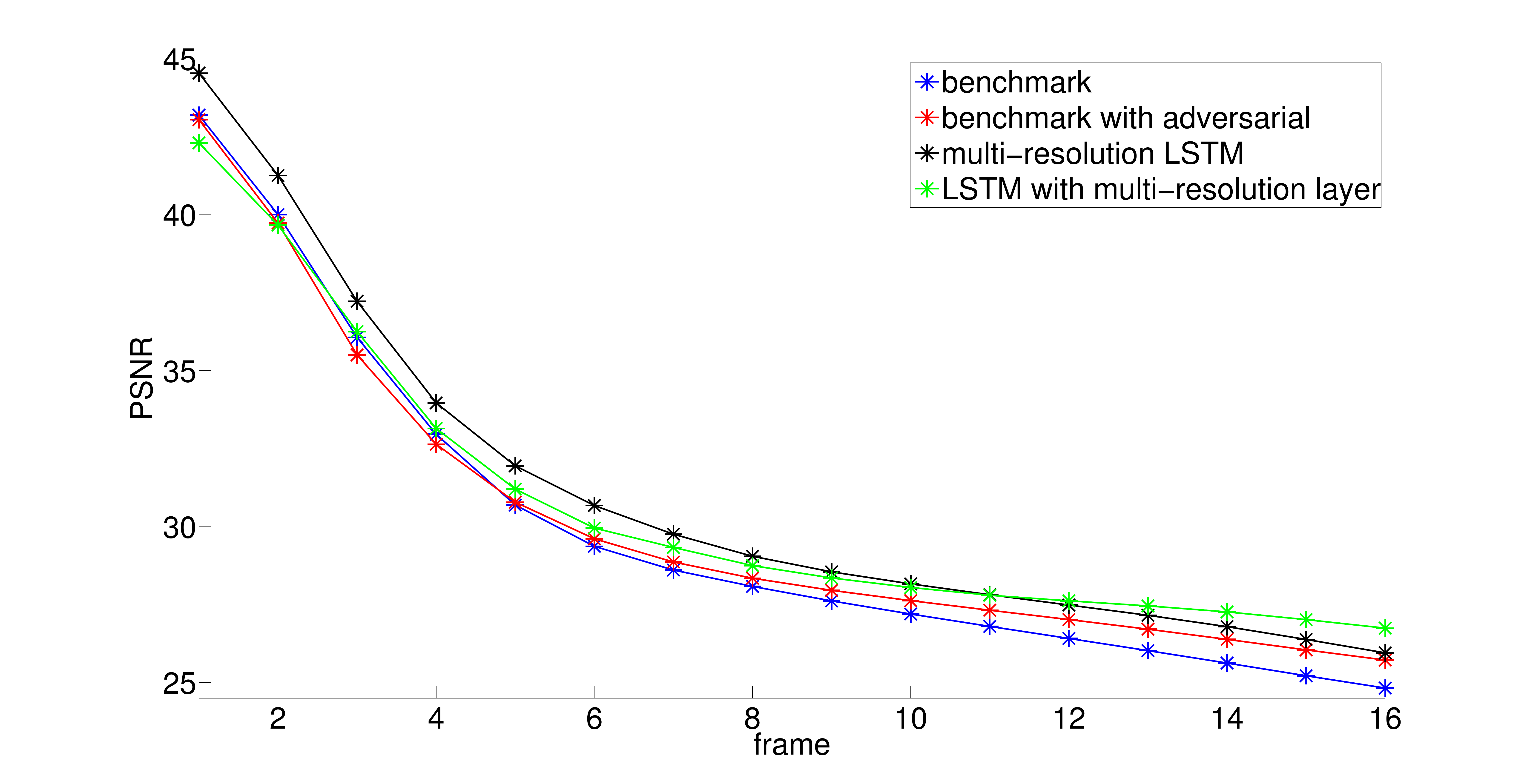}
		\caption{PSNR of predicted frames against prediction time. The benchmark model, benchmark model with adversarial, multi-resolution LSTM, LSTM with multi-resolution layer correspond to models 5,6,7,8 respectively in Tab.~\ref{PSNR_compare}. LSTM with multi-resolution layer has a better long term prediction accuracy compared to other models. The PSNR is obtained by first computing MSE by averaging squared errors over all pixels over all frames and all sequences, and then converting the resulting MSE to PSNR.}
     
\label{error_comparison}
\end{figure} 

For the discriminative model, we use a 3D convolutional neural network with an encoder-decoder structure. The encoder uses three 3D strided convolution layers with all layers using $7 \times 7 \times 7 $ convolutional kernel. The decoder uses three 3D strideded deconvolution layers with all layers with the same receptive field. We use batch normalization \cite{ioffe2015batch} and leaky ReLU \cite{he2015delving} except for the last layer. For multi-resolution network, the discriminative for the higher scale uses only two 3D convolution layer and deconvolution layer as the sequence length is reduced by 2.

We present results for several models. The weights for reconstruction error and prediction error are set to be $\lambda_{rec}=1, \lambda_{pred} =1$ in all models. For models use adversarial training, we set the weight $\lambda_{adv} =0.1$. For multi-resolution LSTM network, the weights are set the same in different scale. The discriminative model and generative model are trained both use Adam algorithm\cite{kingma2014adam} both with learning rate of 0.001 decreasing by a factor of 10 halfway through training. To avoid exploding gradient for generative model, we perform gradient clipping by setting the $l_2$ norm maximum at 0.001. In all adversarial training case, the discriminative model is updated once every two iterations.

During testing stage, the observation sequences lasts 16 frames and the predictor generates 16 future frames based on the observed frames. To evaluate the performance of different approaches we compute the Peak Signal to Noise Ratio (PSNR) between the true future frames $X$ and predicted future frames $Y$. 
\begin{equation*}
PSNR = 10*log_{10}\frac{(max_X^2)}{\frac{1}{N}\sum_{i=1}^N(x_i-y_i)^2}
\label{PSNR_def}
\end{equation*}

\begin{figure*}
        \centering
	
		\includegraphics[width=0.95\textwidth,height = 1.2 \textwidth]{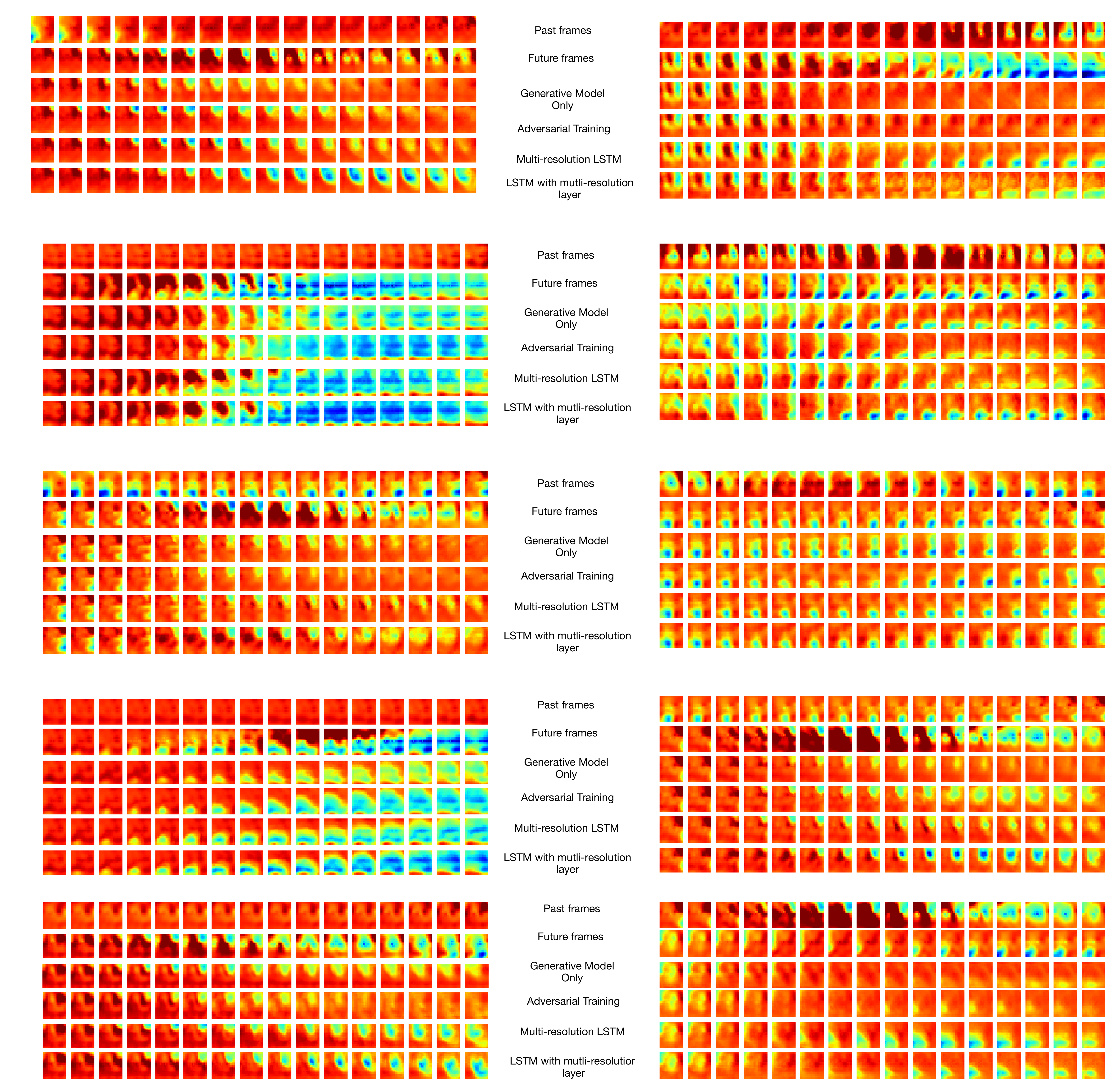}
		\caption{Prediction result comparison between different methods: generative model, adversarial training, multi-resolution LSTM and LSTM with multi-resolution layers correspond to model 5,6,7,8 respectively in Tab.~\ref{PSNR_compare}. }
        
\label{all_sequences}
\end{figure*}

Sample prediction comparison are shown in Tab.~\ref{PSNR_compare}. The adversarial training brings improvement compared to using $l_2$ loss alone. It is interesting to note that even the PSNR is based on $l_2$ metric, adding adversarial training into the loss function gets better prediction accuracy. Discriminative model helps generative model to learn the long term dependencies. The further the prediction the more significant is the accuracy gain from adversarial training, which is shown in Fig.~\ref{error_comparison}. Comparing among all structures using adversarial learning, LSTM with multi-resolution layer and multi-resolution LSTM both have a significant gain compared to the benchmark model. Even though the multi-resolution LSTM achieved more gains for prediction up to 10 frames ahead, the LSTM with multi-resolution layers take over after 10 frames. PSNR increase by using LSTM with multi-resolution layer at $16th$ frames gets as high as 1.92 dB compared to the benchmark model. This is remarkable as the LSTM with multi-resolution layers have about the same number of parameters as the benchmark model. In Fig.~\ref{all_sequences}, we show sample results of different models.


\begin{table*}
\scalebox{0.8}{
\begin{tabular}{  p{2cm} | p{2cm}  p{2cm} p{2cm}  p{2.5 cm} | p{2 cm}  p{2 cm} p{2 cm}  p{2cm}}
\toprule
Generative model & ConvLSTM 64-64 & ConvLSTM 64-64 & Multi-resolution LSTM 64-64, 64-64 & LSTM with multi-resolution layer 64-64  & ConvLSTM 128-128  & ConvLSTM 128-128 & Multi-resolution LSTM 128-128, 128-128  & LSTM with multi-resolution layer 128-128  \\
\hline
Number of parameters in the generative model 
 &  4123266 & 4123266& 4123266 and 8265732  & 4123524 &15619330 & 15619330 & 15619330 and 31277060 & 15619844 \\
\toprule
Discriminative model number of feature maps per layer & None & 32,32,4,32,32 &   32,4,32 and 32,32,4,32,32 &  32,32,4,32,32 & None & 32,32,4,32,32  & 32,4,32 and  32,32,4,32,32 & 32,32,4,32,32  \\
\toprule
PSNR of all frames & 27.8737 & 28.3426 &28.8903 & \textbf{29.0372} & 27.9942 & 28.5317 & 29.0931 & \textbf{29.1741}  \\
\bottomrule
 \end{tabular}}
 \caption{Comparison of the accuracy and number of parameters of all models. 64-64 represents the number of convolution LSTM cells in layer 1 and 2 are both 64. All the convolution LSTM cells uses $5 \times 5$ kernel. The multi-resolution LSTM structure has two scales, each scale has two layer convolution LSTM cells. 64-64, 64-64 means each scale uses a 64-64 two layer LSTM. Model 1 and 5 are trained with $l_2$ loss alone. }
 \label{PSNR_compare}
 \end{table*}

\section{ACKNOWLEDGEMENT}
This work was funded by National Science Foundation award CCF-1422914.
 \vspace{-0.05in}

\section{Conclusion}
In this work, we have proposed two ways to do video prediction using multi-resolution presentations. The first approach uses a novel LSTM structure with multi-resolution layers for long term video prediction.  The network creates a temporal highway in the upper-layer to capture the long-term dependencies between video frames. 
The second approach uses two scale multi-resolution LSTM. We compare the performance of these two approaches against single resolution benchmark model and demonstrate the advantage of using multi-resolution representation of LSTM. Both multi-resolution LSTM and LSTM with multi-resolution layers have better performance than single resolution representation  when they all use adversarial training. The long term prediction accuracy using LSTM with multi-resolution layers are much higher than the benchmark models with similar number of parameters. We also demonstrate that all models benefit from energy-based adversarial training which is accomplished by using a 3D CNN based encoder-decoder structure.

{\small
\bibliographystyle{ieee}
\bibliography{mybib}

\begin{thebibliography}{10}\itemsep=-1pt

\bibitem{acharya2011automatic}
U.~R. Acharya, S.~V. Sree, and J.~S. Suri.
\newblock Automatic detection of epileptic eeg signals using higher order
  cumulant features.
\newblock {\em International journal of neural systems}, 21(05):403--414, 2011.

\bibitem{arjovsky2017wasserstein}
M.~Arjovsky, S.~Chintala, and L.~Bottou.
\newblock Wasserstein gan.
\newblock {\em arXiv preprint arXiv:1701.07875}, 2017.

\bibitem{bahdanau2014neural}
D.~Bahdanau, K.~Cho, and Y.~Bengio.
\newblock Neural machine translation by jointly learning to align and
  translate.
\newblock {\em arXiv preprint arXiv:1409.0473}, 2014.

\bibitem{bandarabadi2015epileptic}
M.~Bandarabadi, C.~A. Teixeira, J.~Rasekhi, and A.~Dourado.
\newblock Epileptic seizure prediction using relative spectral power features.
\newblock {\em Clinical Neurophysiology}, 126(2):237--248, 2015.

\bibitem{cho2014learning}
K.~Cho, B.~Van~Merri{\"e}nboer, C.~Gulcehre, D.~Bahdanau, F.~Bougares,
  H.~Schwenk, and Y.~Bengio.
\newblock Learning phrase representations using rnn encoder-decoder for
  statistical machine translation.
\newblock {\em arXiv preprint arXiv:1406.1078}, 2014.

\bibitem{chua2009automatic}
K.~Chua, V.~Chandran, U.~Acharya, and C.~Lim.
\newblock Automatic identification of epileptic electroencephalography signals
  using higher-order spectra.
\newblock {\em Proceedings of the Institution of Mechanical Engineers, Part H:
  Journal of Engineering in Medicine}, 223(4):485--495, 2009.

\bibitem{eftekhar2014ngram}
A.~Eftekhar, W.~Juffali, J.~El-Imad, T.~G. Constandinou, and C.~Toumazou.
\newblock Ngram-derived pattern recognition for the detection and prediction of
  epileptic seizures.
\newblock {\em PloS one}, 9(6):e96235, 2014.

\bibitem{gadhoumi2012discriminating}
K.~Gadhoumi, J.-M. Lina, and J.~Gotman.
\newblock Discriminating preictal and interictal states in patients with
  temporal lobe epilepsy using wavelet analysis of intracerebral eeg.
\newblock {\em Clinical neurophysiology}, 123(10):1906--1916, 2012.

\bibitem{goodfellow2014generative}
I.~Goodfellow, J.~Pouget-Abadie, M.~Mirza, B.~Xu, D.~Warde-Farley, S.~Ozair,
  A.~Courville, and Y.~Bengio.
\newblock Generative adversarial nets.
\newblock In {\em Advances in neural information processing systems}, pages
  2672--2680, 2014.

\bibitem{graves2014neural}
A.~Graves, G.~Wayne, and I.~Danihelka.
\newblock Neural turing machines.
\newblock {\em arXiv preprint arXiv:1410.5401}, 2014.

\bibitem{gulcehre2017memory}
C.~Gulcehre, S.~Chandar, and Y.~Bengio.
\newblock Memory augmented neural networks with wormhole connections.
\newblock {\em arXiv preprint arXiv:1701.08718}, 2017.

\bibitem{he2015delving}
K.~He, X.~Zhang, S.~Ren, and J.~Sun.
\newblock Delving deep into rectifiers: Surpassing human-level performance on
  imagenet classification.
\newblock In {\em Proceedings of the IEEE international conference on computer
  vision}, pages 1026--1034, 2015.

\bibitem{he2016deep}
K.~He, X.~Zhang, S.~Ren, and J.~Sun.
\newblock Deep residual learning for image recognition.
\newblock In {\em Proceedings of the IEEE Conference on Computer Vision and
  Pattern Recognition}, pages 770--778, 2016.

\bibitem{hochreiter1997long}
S.~Hochreiter and J.~Schmidhuber.
\newblock Long short-term memory.
\newblock {\em Neural computation}, 9(8):1735--1780, 1997.

\bibitem{ioffe2015batch}
S.~Ioffe and C.~Szegedy.
\newblock Batch normalization: Accelerating deep network training by reducing
  internal covariate shift.
\newblock {\em arXiv preprint arXiv:1502.03167}, 2015.

\bibitem{kaiser2015neural}
{\L}.~Kaiser and I.~Sutskever.
\newblock Neural gpus learn algorithms.
\newblock {\em arXiv preprint arXiv:1511.08228}, 2015.

\bibitem{kingma2014adam}
D.~Kingma and J.~Ba.
\newblock Adam: A method for stochastic optimization.
\newblock {\em arXiv preprint arXiv:1412.6980}, 2014.

\bibitem{ledig2016photo}
C.~Ledig, L.~Theis, F.~Husz{\'a}r, J.~Caballero, A.~Cunningham, A.~Acosta,
  A.~Aitken, A.~Tejani, J.~Totz, Z.~Wang, et~al.
\newblock Photo-realistic single image super-resolution using a generative
  adversarial network.
\newblock {\em arXiv preprint arXiv:1609.04802}, 2016.

\bibitem{li2013seizure}
S.~Li, W.~Zhou, Q.~Yuan, and Y.~Liu.
\newblock Seizure prediction using spike rate of intracranial eeg.
\newblock {\em IEEE transactions on neural systems and rehabilitation
  engineering}, 21(6):880--886, 2013.

\bibitem{lotter2015unsupervised}
W.~Lotter, G.~Kreiman, and D.~Cox.
\newblock Unsupervised learning of visual structure using predictive generative
  networks.
\newblock {\em arXiv preprint arXiv:1511.06380}, 2015.

\bibitem{mathieu2015deep}
M.~Mathieu, C.~Couprie, and Y.~LeCun.
\newblock Deep multi-scale video prediction beyond mean square error.
\newblock {\em arXiv preprint arXiv:1511.05440}, 2015.

\bibitem{netoff2009seizure}
T.~Netoff, Y.~Park, and K.~Parhi.
\newblock Seizure prediction using cost-sensitive support vector machine.
\newblock In {\em 2009 Annual International Conference of the IEEE Engineering
  in Medicine and Biology Society}, pages 3322--3325. IEEE, 2009.

\bibitem{oord2016pixel}
A.~v.~d. Oord, N.~Kalchbrenner, and K.~Kavukcuoglu.
\newblock Pixel recurrent neural networks.
\newblock {\em arXiv preprint arXiv:1601.06759}, 2016.

\bibitem{radford2015unsupervised}
A.~Radford, L.~Metz, and S.~Chintala.
\newblock Unsupervised representation learning with deep convolutional
  generative adversarial networks.
\newblock {\em arXiv preprint arXiv:1511.06434}, 2015.

\bibitem{rocktaschel2015reasoning}
T.~Rockt{\"a}schel, E.~Grefenstette, K.~M. Hermann, T.~Ko{\v{c}}isk{\`y}, and
  P.~Blunsom.
\newblock Reasoning about entailment with neural attention.
\newblock {\em arXiv preprint arXiv:1509.06664}, 2015.

\bibitem{salimans2016improved}
T.~Salimans, I.~Goodfellow, W.~Zaremba, V.~Cheung, A.~Radford, and X.~Chen.
\newblock Improved techniques for training gans.
\newblock In {\em Advances in Neural Information Processing Systems}, pages
  2226--2234, 2016.

\bibitem{santoro2016one}
A.~Santoro, S.~Bartunov, M.~Botvinick, D.~Wierstra, and T.~Lillicrap.
\newblock One-shot learning with memory-augmented neural networks.
\newblock {\em arXiv preprint arXiv:1605.06065}, 2016.

\bibitem{sorensen2010automatic}
T.~L. Sorensen, U.~L. Olsen, I.~Conradsen, J.~Duun-Henriksen, T.~W. Kjaer,
  C.~E. Thomsen, and H.~B.~D. S{\o}rensen.
\newblock Automatic epileptic seizure onset detection using matching pursuit.
\newblock 2010.

\bibitem{srivastava2015unsupervised}
N.~Srivastava, E.~Mansimov, and R.~Salakhutdinov.
\newblock Unsupervised learning of video representations using lstms.
\newblock {\em CoRR, abs/1502.04681}, 2, 2015.

\bibitem{srivastava2015highway}
R.~K. Srivastava, K.~Greff, and J.~Schmidhuber.
\newblock Highway networks.
\newblock {\em arXiv preprint arXiv:1505.00387}, 2015.

\bibitem{taigman2016unsupervised}
Y.~Taigman, A.~Polyak, and L.~Wolf.
\newblock Unsupervised cross-domain image generation.
\newblock {\em arXiv preprint arXiv:1611.02200}, 2016.

\bibitem{temko2011eeg}
A.~Temko, E.~Thomas, W.~Marnane, G.~Lightbody, and G.~Boylan.
\newblock Eeg-based neonatal seizure detection with support vector machines.
\newblock {\em Clinical Neurophysiology}, 122(3):464--473, 2011.

\bibitem{trischler2016natural}
A.~Trischler, Z.~Ye, X.~Yuan, and K.~Suleman.
\newblock Natural language comprehension with the epireader.
\newblock {\em arXiv preprint arXiv:1606.02270}, 2016.

\bibitem{van2016wavenet}
A.~van~den Oord, S.~Dieleman, H.~Zen, K.~Simonyan, O.~Vinyals, A.~Graves,
  N.~Kalchbrenner, A.~Senior, and K.~Kavukcuoglu.
\newblock Wavenet: A generative model for raw audio.
\newblock {\em CoRR abs/1609.03499}, 2016.

\bibitem{viventi2011flexible}
J.~Viventi, D.-H. Kim, L.~Vigeland, E.~S. Frechette, J.~A. Blanco, Y.-S. Kim,
  A.~E. Avrin, V.~R. Tiruvadi, S.-W. Hwang, A.~C. Vanleer, et~al.
\newblock Flexible, foldable, actively multiplexed, high-density electrode
  array for mapping brain activity in vivo.
\newblock {\em Nature neuroscience}, 14(12):1599--1605, 2011.

\bibitem{xingjian2015convolutional}
S.~Xingjian, Z.~Chen, H.~Wang, D.-Y. Yeung, W.-k. Wong, and W.-c. Woo.
\newblock Convolutional lstm network: A machine learning approach for
  precipitation nowcasting.
\newblock In {\em Advances in Neural Information Processing Systems}, pages
  802--810, 2015.

\bibitem{zhao2016energy}
J.~Zhao, M.~Mathieu, and Y.~LeCun.
\newblock Energy-based generative adversarial network.
\newblock {\em arXiv preprint arXiv:1609.03126}, 2016.

\end{thebibliography}
}

\end{document}